\documentclass[conference]{IEEEtran}
\IEEEoverridecommandlockouts

\usepackage{cite}
\usepackage{amsmath,amssymb,amsfonts}
\usepackage{algorithmic}
\usepackage{graphicx}
\usepackage{textcomp}
\usepackage{xcolor}
\usepackage{diagbox}
\usepackage{multirow}
\usepackage{pifont} 
\usepackage{booktabs}

\newcommand{\cmark}{\ding{51}} 
\newcommand{\xmark}{\ding{55}} 

\def\BibTeX{{\rm B\kern-.05em{\sc i\kern-.025em b}\kern-.08em
    T\kern-.1667em\lower.7ex\hbox{E}\kern-.125emX}}

\makeatletter
\newcommand{\linebreakand}{
  \end{@IEEEauthorhalign}
  \hfill\mbox{}\par
  \mbox{}\hfill\begin{@IEEEauthorhalign}
}
\makeatother

\begin{document}

\title{AntennaFlow: A Generative Flow Model for Offset Correction in Phaseless Antenna Testing
}

\author{
\IEEEauthorblockN{1\textsuperscript{st} Yongzhi Li}
\IEEEauthorblockA{\textit{College of Computing and Data Science} \\
\textit{Nanyang Technological University}\\
Singapore \\
YONGZHI001@e.ntu.edu.sg}
\and
\IEEEauthorblockN{2\textsuperscript{nd} Chongting Shen}
\IEEEauthorblockA{\textit{Beihang Sino-French Engineer School} \\
\textit{Beihang University}\\
Beijing, China \\
schongting@buaa.edu.cn}
\and
\IEEEauthorblockN{3\textsuperscript{rd} Menglin Chen}
\IEEEauthorblockA{\textit{Beihang Sino-French Engineer School} \\
\textit{Beihang University}\\
Beijing, China \\
elvin\_chen@buaa.edu.cn}
\and
\linebreakand 
\IEEEauthorblockN{4\textsuperscript{th} Xun Jiang}
\IEEEauthorblockA{\textit{School of Computer Science and Engineering}\\
\textit{University of Electronic Science and Technology of China}\\
Chengdu, China \\
xun\_jiang@std.uestc.edu.cn}
\and 
\IEEEauthorblockN{5\textsuperscript{th} Zhengpeng Wang}
\IEEEauthorblockA{\textit{Electronic Information Engineering} \\
\textit{Beihang University}\\
Beijing, China \\
wangzp@buaa.edu.cn}
}

\maketitle

\begin{abstract}
Near-field to far-field transformation is central to large-aperture antenna testing, yet two coupled challenges remain: costly phase acquisition at millimeter-wave bands and violations of the centering assumption under offset mounting. Existing methods address these issues separately, requiring either dense full-field data or offset vectors. We tackle both jointly by exploiting a key observation: amplitude fields under different offsets are coordinate-transformed views of the same near field. The challenge is to recover the center-aligned field from offset amplitudes without a phase or offset vector. We propose AntennaFlow, a three-stage framework: a contrastively learned encoder that maps offset views to an offset-invariant embedding, a deterministic flow-matching transport that maps offset amplitudes to center-aligned ones, and the Simplified Extrapolation Technique, whose Green-function Taylor expansion is valid only for centered fields. Experiments show that AntennaFlow enables fast, phaseless, offset-vector-free NF--FF reconstruction from sparse amplitude-only measurements, consistently outperforming existing baselines while preserving physical consistency. 
\end{abstract}


\section{Introduction}
With the rise of 6G~\cite{jiang2021road}, large-aperture arrays such as massive MIMO and phased arrays~\cite{bjornson2019massive} are widely used in satellites and radar. Large electrical size leads to prohibitively long Rayleigh distances, making direct far-field testing impractical. Near-field to far-field (NF--FF) transformation~\cite{bucci2005near} thus serves as the primary Over-The-Air (OTA) testing approach, enabling accurate measurements via near-field.


Despite its maturity, near-field measurement still faces two key challenges in practice: phase acquisition and offset placements~\cite{sorensen2012estimate}. At millimeter-wave frequencies, accurate phase measurement requires stable RF cables and precise synchronization, leading to high cost and long acquisition time. To address this, phaseless methods such as the Extrapolation Technique (ET)~\cite{1140519} and Simplified Extrapolation Technique (SET)~\cite{yu2025antenna} reconstruct the far field from amplitude-only data, avoiding the burden of phase measurements.

\begin{figure}[tbp]
    \centering
    \includegraphics[width=0.85\linewidth]{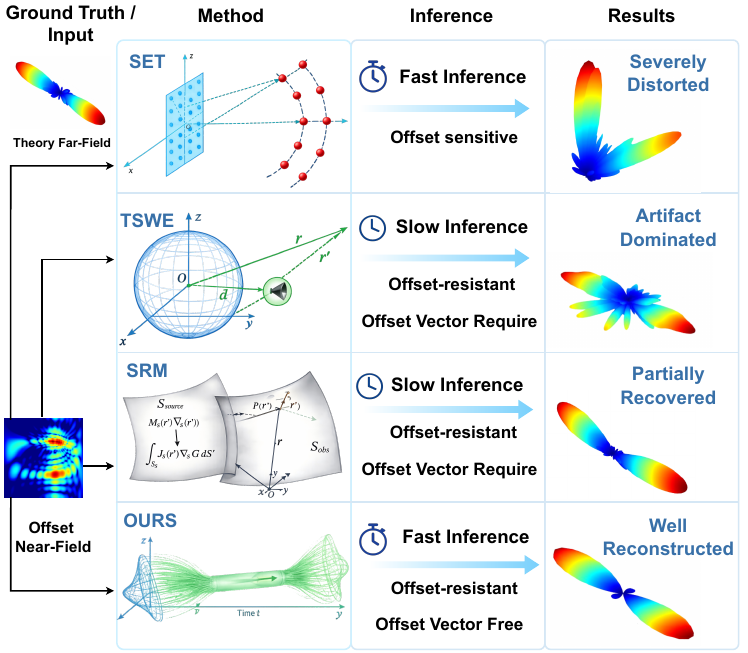} 
    \caption{Comparison of far-field reconstruction under offset mounting: prior methods suffer from offset sensitivity or reliance on offset vectors, while AntennaFlow achieves superior reconstruction quality.}
    \label{fig:fig1}
\end{figure}

However, array antennas are often mounted as subsystems on complex platforms, including satellites, vehicles, and radar domes, making precise alignment with the measurement coordinate center difficult and degrading the measured amplitude and phase. TSWE~\cite{cornelius2017spherical} compensates for offset mounting, while SRM~\cite{8000577} recovers phase and can also mitigate offset. However, both suffer from low sampling efficiency and reliance on offset vectors, limiting practical applicability. Consequently, few methods can efficiently handle offset correction under phaseless settings, motivating offset-vector-free offset correction, as illustrated in Fig.~\ref{fig:fig1}.


We resolve this by noting that different offset placements yield only coordinate-transformed views of the same near field. The challenge is to recover the centered field from offset amplitudes without phase and offset vector. Based on this, AntennaFlow uses contrastive learning to obtain an offset-invariant embedding, a deterministic flow to map offset amplitudes to centered ones, and SET to extract the far field from the centered output. The three stages thus remove offset-induced errors in the NF-FF pipeline without ever observing the offset vector or any phase information.
The main contributions of this work are summarized as follows:

\begin{itemize}
    \item We unify phaseless NF-FF transformation and offset correction into an end-to-end amplitude calibration without phase or offset vectors.

    \item We realize this as AntennaFlow: offset-invariant amplitude embedding, deterministic flow-matching transport, and center-aligned SET extraction for efficient NF-FF reconstruction without phase or offset vectors.

    \item To our knowledge, this is the first application of generative AI to NF--FF transformation, achieving consistent gains in reconstruction quality, efficiency, and robustness under realistic measurement perturbations.
\end{itemize}

\section{Background}
\subsection{Antenna Testing}
\subsubsection{Phaseless Testing} Spherical near-field measurement, based on Hansen’s spherical wave expansion (SWE)~\cite{hansen1988spherical}, is the benchmark for high-accuracy antenna testing. However, costly and unstable phase measurements have driven phaseless approaches that reconstruct the far field from amplitude-only data. SRM~\cite{8000577} estimates equivalent sources via iterative optimization, while ET~\cite{1140519} applies polynomial fitting for gain calibration. In our work, we adopt the Simplified Extrapolation Technique (SET)~\cite{yu2025antenna} as our core solver, which directly fits amplitude-decay curves to avoid iterative local minima and slow inference, and naturally aligns with our calibration through its dependence on array centering.

\subsubsection{Offset Correction} Offset mounting enlarges the enclosing sphere, increasing Nyquist sampling demands, cost, and measurement degradation. TSWE \cite{cornelius2017spherical} requires full-field data, SRM relies on costly phase retrieval, and SET, derived from Green’s formula with a local Taylor expansion, cannot handle offsets. Existing methods thus require either an offset vector and complete data or heavy computation. In contrast, we exploit shared structures across offset-induced distortions and learn a calibration map for correction under sparse, phaseless measurements without offset vector inputs. Table I summarizes the comparison.


\begin{table}[t]
  \centering
  \caption{Comparison of the previous methods and the proposed method}
  \label{tab:method_comparison}
  \begin{tabular}{c|cccccc}
    \toprule
    \multirow{2}{*}{Capabilities} & \multicolumn{6}{c}{Method} \\
    \cmidrule(lr){2-7}
     & ET & SET  & SWE  & TSWE& SRM & Ours \\
    \midrule
    Sparse Sampling              &\xmark&\cmark & \xmark & \xmark & \xmark & \cmark\\
    Fast Inference               &\xmark&\cmark & \xmark & \xmark & \xmark & \cmark \\
    Phaseless &\cmark &\cmark &\xmark&\xmark &\cmark&\cmark \\
    Offset Correction             &\xmark&\xmark & \xmark & \cmark & \cmark & \cmark \\
    Offset-Vector Free    & --  & --&  -- &\xmark& \xmark & \cmark \\
    \bottomrule
  \end{tabular}
\end{table}

\subsection{Generative AI}
\subsubsection{Image Generation with Diffusion and Flow Models}Recent diffusion-based generative models \cite{ho2020denoising, lipman2022flow} have achieved strong performance in image generation \cite{li2025spotdiff}. Our task, however, maps offset near-field amplitudes to physically meaningful center-aligned counterparts, rather than generating diverse images. SDE-based models introduce stochasticity that benefits natural image generation but leads to undesirable fluctuations for precise near-field calibration. 
In contrast, ODE-based flow-matching models provide deterministic mapping, making it better aligned with the requirements of this task.

\subsubsection{Contrastive Learning} 
Contrastive learning organizes representations by pulling positive pairs together and pushing negative pairs apart.
In our setting, we use antenna identity as the supervision signal in a supervised contrastive framework~\cite{khosla2020supervised}, treating different offset views of the same antenna as positives, while views from different antennas are treated as negatives.
This encourages the encoder to capture features that are invariant to spatial offsets, resulting in robust near-field representations that support offset-vector-free reconstruction.

\begin{figure*}[htbp]
    \centering
    \includegraphics[width=0.85\linewidth]{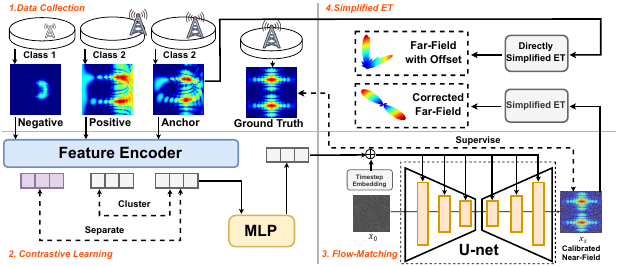}
    \caption{The AntennaFlow pipeline for NF-FF transformation. (1) Data Collection: Near-field data are sampled from different antenna configurations. (2) A feature encoder is trained using a supervised contrastive learning framework to learn discriminative representations. (3) Flow Matching Transport: A U-Net-based architecture is used to learn the flow-matching velocity field for center near-field generation. (4) Far-Field Reconstruction via SET. Note that SET requires measurements from multiple distinct surfaces; for simplicity, only one surface is illustrated in the figure.}
    \label{fig:pipeline_fig}
\end{figure*}

\section{Method}

\subsection{Design Overview}

AntennaFlow is motivated by a simple measurement observation: offset placements do not alter the underlying physical near field, but only its coordinate representation. The core challenge is therefore to recover a canonical, center-aligned field from amplitude-only measurements without access to phase or offset vectors.
To address this, we decompose the problem into three components. First, we learn an offset-invariant representation $\mathbf{f}$ that captures antenna-specific characteristics shared across different offset measurements. Second, we model a deterministic conditional transport that maps an input amplitude field to its center-aligned counterpart, with $\mathbf{f}$ providing invariant structural guidance. Third, we ensure that the reconstructed field satisfies the assumptions required by the downstream phaseless solver (SET), whose formulation is valid only under center alignment.
Together, these components enable an end-to-end pipeline for offset correction and phaseless NF–FF transformation, as illustrated in Fig.~\ref{fig:pipeline_fig}.

\subsection{AntennaFlow for Near-Field Calibration}

AntennaFlow uses a contrastively pretrained ResNet encoder~\cite{he2016deep} and a U-Net flow-matching backbone~\cite{ronneberger2015u} conditioned on its embedding to learn deterministic transport from offset amplitudes to center-aligned amplitudes.

\subsubsection{Offset-Invariant Antenna Embedding}

For a batch of labeled \emph{pretraining} samples $\{(X_i, Y_i)\}_{i=1}^{B}$, $X_i$ denotes a near-field amplitude and $Y_i$ denotes the antenna identity label, which is available only during the training stage; the encoded representation is
$h_i = E(X_i)$,
where $E(\cdot)$ is the feature encoder. With this label assignment, the supervised contrastive paradigm~\cite{khosla2020supervised} becomes a direct mechanism for compressing the set of offset views of each antenna into a tight cluster: for each anchor $i$, its positive set
$P(i) = \{\, j \neq i \mid Y_j = Y_i \,\}$
collects offset views of the same antenna, while the remaining $Y_j \neq Y_i$ are negatives.
The cosine similarity scaled by temperature $\tau$ is computed as:
\vspace{-3pt}
\begin{equation}
    s_{ij} = \frac{h_i^\top h_j}{\tau}.
\end{equation}

The supervised InfoNCE loss is defined as:
\begin{equation}
    \mathcal{L}_{\text{SupCon}}
    =
    - \frac{1}{B}
    \sum_{i=1}^{B}
    \frac{1}{|P(i)|}
    \sum_{p \in P(i)}
    \log
    \frac{
        \exp(s_{ip})
    }{
        \sum\limits_{\substack{j=1,j \neq i}}^{B}
        \exp(s_{ij})
    }.
\end{equation}

Minimizing $\mathcal{L}_{\text{SupCon}}$ encourages approximate invariance of the embedding across different offset near-fields. A learnable MLP $M(\cdot)$ then produces the final conditional embedding:
\begin{equation}
    \mathbf{f} = M\big(E(X_i)\big),
\end{equation}
which supports the next deterministic transport. 
\subsubsection{Deterministic Transport via Conditional Flow Matching}

Given an offset amplitude $x$ and its centered counterpart $x_1$, we learn a conditional transport rather than an unconditional generation. Specifically, we construct a displacement interpolation between a noise prior $x_0 \sim \mathcal{N}(0, I)$ and $x_1$ as $x_t = (1 - t)x_0 + t x_1$, whose time derivative yields a constant ground-truth velocity $v^\ast(x_t) = \frac{d x_t}{dt} = x_1 - x_0$.
The velocity field $v_\theta(\cdot,\cdot,\mathbf{f})$ is parameterized by a U-Net that takes the interpolated state $x_t$, the timestep $t$, and the conditional embedding $\mathbf{f}$ as inputs, and outputs a velocity vector of the same dimensionality as $x_t$. The model is trained via:
\begin{equation}
\mathcal{L}_{\text{FM}} =
\mathbb{E}_{x_0, x_1, t}
\left\|
v_\theta(x_t, t, \mathbf{f}) - v^\ast(x_t)
\right\|_2^2.
\end{equation}

Crucially, the conditioning $\mathbf{f}=M(E(x))$ is extracted from the input amplitude $x$, serving as an offset-invariant anchor of the antenna identity. As a result, the transport is conditioned on $x$ through $\mathbf{f}$, rather than discarding input information.

At inference, we integrate from $x_0 \sim \mathcal{N}(0,I)$ under $\mathbf{f}$ using the U-Net velocity field:
\begin{equation}
x_{t_{k+1}} = x_{t_k} + \Delta t\, v_\theta(x_{t_k}, t_k, \mathbf{f}),
\end{equation}
and the final state $\hat{x}_1$ is taken as the generated sample.

\subsection{Far-Field Transformation via SET}

After offset removal, we apply SET~\cite{yu2025antenna} as the NF-FF transformation module. 
Specifically, AntennaFlow generates calibrated near-field amplitudes conditioned on the offset near-field amplitudes of multiple measurement radii. 
In this work, we generate 3 calibrated near-field maps at distinct distances $r_1$, $r_2$, and $r_3$, which serve as the required inputs for SET.

Let $\hat{x}_1(r,\theta,\phi)$ denote the calibrated near-field amplitude at radius $r$. 
Under the center-alignment property guaranteed by AntennaFlow, the spatial variation of the near-field is parameterized by angular coordinates $(\theta,\phi)$. 
Following the SET formulation, we model the distance-dependent power decay by approximating the product of the power density $P(r,\theta,\phi)=|\hat{x}_1(r,\theta,\phi)|^2$ and $r^2$ using a truncated polynomial expansion in $r^{-2k}$, derived from the Taylor expansion of the Green’s function:
\begin{equation} 
\label{eq:extrapolation_poly} 
P(r, \theta, \phi) \cdot r^2 \approx \sum_{k=0}^{K} A'_{2k}(\theta, \phi) \cdot r^{-2k}.
\end{equation}

where $A'_{00}$ is the zero-order intercept coefficient representing the far-field radiation characteristic, while higher-order terms account for the reactive and radiative near-field decay components. In practice, for sufficiently large measurement distances, higher-order terms decay rapidly with increasing powers of $r^{-2}$, and prior work has shown that accurate approximation can be achieved using only the first three terms:
\[
P(r, \theta, \phi) \cdot r^2 \approx A'_{00}(\theta,\phi) + \frac{A'_{02}(\theta,\phi)}{r^2} + \frac{A'_{04}(\theta,\phi)}{r^4}.
\]

To estimate the far-field coefficient $A'_{00}$, AntennaFlow generates
calibrated amplitude maps at three distinct mid-field distances, $r_1$,
$r_2$, and $r_3$. For each distance $r_m$, we define the processed observation as:
\begin{equation}
y_m(\theta, \phi) = |\hat{x}_1(r_m,\theta, \phi)|^2 r_m^2 = P(r_m, \theta, \phi) \cdot r_m^2. 
\end{equation}
The resulting observations provide a
linear system for solving the polynomial coefficients:

\begin{equation}
\label{eq:matrix_system}
\begin{bmatrix}
y_1(\theta, \phi) \\
y_2(\theta, \phi) \\
y_3(\theta, \phi)
\end{bmatrix}
=
\begin{bmatrix}
1 & r_1^{-2} & r_1^{-4}\\
1 & r_2^{-2} & r_2^{-4}\\
1 & r_3^{-2} & r_3^{-4}
\end{bmatrix}
\begin{bmatrix}
A'_{00}(\theta, \phi) \\
A'_{02}(\theta, \phi) \\
A'_{04}(\theta, \phi)
\end{bmatrix}.
\end{equation}

Solving this system independently for each angular pixel
$(\theta,\phi)$ yields the far-field coefficient $A'_{00}$. The far-field
pattern $F(\theta,\phi)$ is then obtained from this extracted
intercept term:

\begin{equation}
\label{eq:far_field_final}
F(\theta, \phi) \propto \sqrt{A'_{00}(\theta, \phi)}.
\end{equation}

This pipeline enables phaseless, offset-free NF–FF transformation via learned center alignment consistent with SET.

 \subsection{Training Data Collection}
 We construct a synthetic dataset to model near-field amplitude responses of antenna arrays under spatial misalignment. For each antenna, the near field is uniquely determined by its array structure and element parameters, and is computed as
\begin{equation}
E(\mathbf{r}) = \sum_{n=1}^{N} I_n f_n(\theta_n,\phi_n)
\frac{e^{-jk|\mathbf{r}-(\mathbf{r}_n+\mathbf{d})|}}
{|\mathbf{r}-(\mathbf{r}_n+\mathbf{d})|},
\end{equation}
where $N$ denotes the number of elements, $I_n$ the fixed element excitation, $f_n(\theta,\phi)$ the normalized element pattern, $\mathbf{r}_n$ the element position with respect to the measurement center, and $\mathbf{d}$ the offset vector representing array misalignment.

According to prior work on near-field synthesis of nonuniformly spaced arrays~\cite{1144001}, the near-field pattern is governed by the element locations, spacings, and element parameters; consequently, any change in array position or spacing leads to a different near-field distribution. Therefore, under the sampling conditions and parameter space defined in this work, each amplitude map sampled over the spherical domain is generated from a unique antenna configuration and can be treated as a single-solution supervision target for both the generative model and subsequent measurement validation.

\begin{figure}[htbp] 
    \centering

    \includegraphics[width=1.0\linewidth]{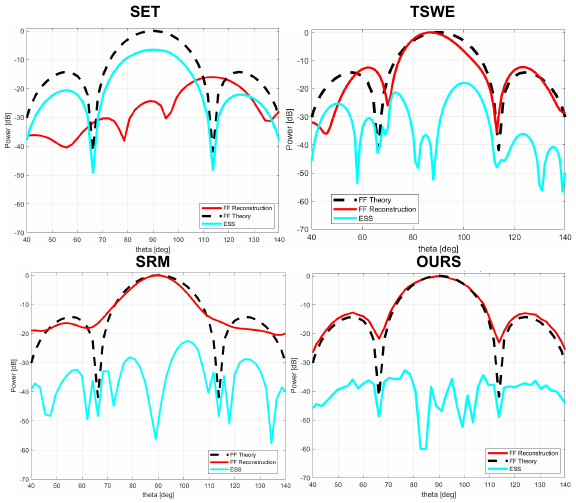} 
    \caption{Comparison of antenna far-field reconstruction using different phaseless reconstruction methods. The red solid curves denote the reconstructed far-field patterns, while the black dashed curves represent the theoretical far-field reference. The cyan curves indicate the corresponding ESS.}
    \label{fig:linechart}
\end{figure}

\begin{figure*}[htbp] 
    \centering

    \includegraphics[width=1.0\linewidth]{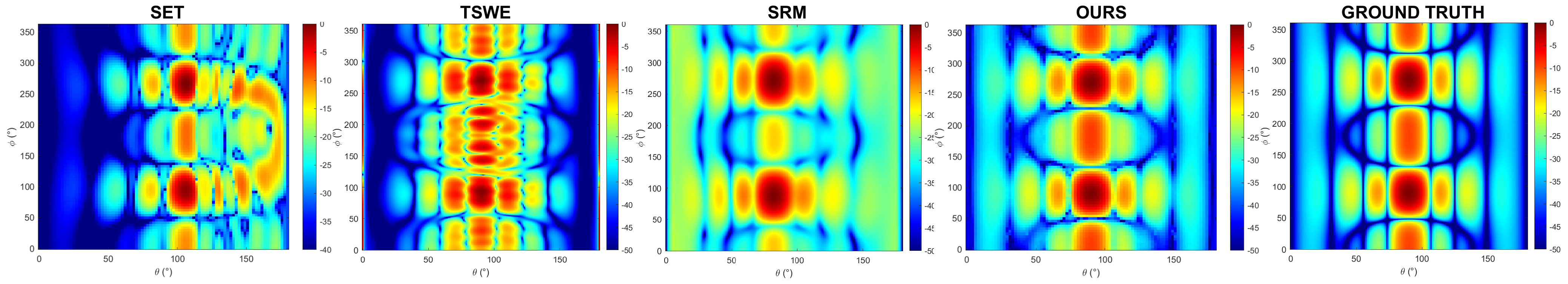} 
    \caption{Visual comparison of reconstructed far-field amplitude patterns under offset conditions. From left to right: SET, TSWE, SRM, AntennaFlow, and the ground truth. AntennaFlow demonstrates the highest fidelity in restoring detailed pattern textures compared to the ground truth.}
    \label{fig:ff-r}
\end{figure*}

\begin{figure}[htbp] 
    \centering

    \includegraphics[width=1.0\linewidth]{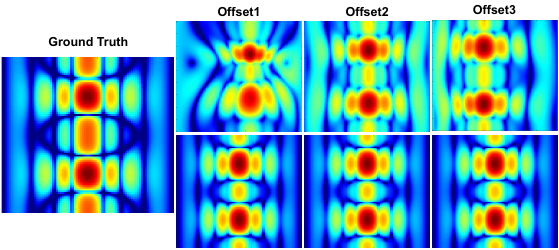} 
    \caption{Reconstruction from offset near-field measurements. Different offset inputs (top) are mapped to consistent center-aligned fields (bottom).}
    \label{fig:2d_vis}
\end{figure}

\section{Experiments and Results}
\subsection{Experiments setup}
\subsubsection{Test Dataset}

To evaluate generalization to unseen antennas, we construct an independent test set of 1200 MATLAB-derived and 300 full-wave FEKO near-fields. The training and test antennas are strictly disjoint and differ in element type, spacing, and aperture size, with additional unseen offset configurations. The FEKO datasets introduce richer electromagnetic effects and a solver shift to assess cross-solver robustness. Together, these settings enable a comprehensive evaluation of reconstruction quality, physical fidelity, and robustness on antennas entirely unseen by the model.

\subsubsection{Evaluation Metrics}

To quantitatively evaluate the reconstructed amplitude patterns in both the far field and the calibrated near-field produced by AntennaFlow, we adopt the Equivalent Stray Signal (ESS) \cite{hess2011historical}, a standard metric in antenna measurements, which interprets reconstruction error as an equivalent stray radiation field superimposed on the target pattern. Let $E_{rec}(\theta)$ and $E_{ref}(\theta)$ denote the normalized amplitudes of the reconstructed and reference patterns, defined either at a fixed measurement radius in the near field or in the far field. Following \cite{yu2025antenna}, the ESS at angle $\theta$ is defined as:

\begin{equation} \label{eq:ess_def} \begin{split} \text{ESS}(\theta) &= 20 \log_{10}(E_{ref}(\theta)) \\ &\quad + 20 \log_{10} \left[ \frac{1 - 10^{-\Delta_{\text{dB}}(\theta)/20}}{2} \right]. \end{split} \end{equation}

\noindent where $\Delta_{\text{dB}}(\theta) = 20 \log_{10}(E_{rec}(\theta)) - 20 \log_{10}(E_{ref}(\theta))$ denotes the pattern difference in decibels. The ESS is evaluated using its Root Mean Square (RMS) and Peak values: a lower \textit{RMS ESS} indicates better overall reconstruction fidelity, while \textit{Peak ESS} reflects the maximum local deviation. By design, ESS is more sensitive to errors in high-energy regions and less sensitive to discrepancies in low-energy areas with limited physical relevance, thereby emphasizing deviations that impact radiation characteristics and better reflecting physical field behavior rather than local numerical artifacts. Consequently, it effectively characterizes the influence of reconstruction errors on sidelobe levels and other key features. 



\subsubsection{Compared Methods}
To rigorously validate the proposed framework, we compare it with SET, TSWE, and SRM, evaluating near-field and far-field reconstruction quality.


\subsection{Results}

\subsubsection{Quantitative Analysis}

As shown in Fig.~\ref{fig:linechart}, existing methods degrade under offset conditions: SET produces pattern shifts, TSWE fails to recover the main lobe, and SRM shows sidelobe distortions. In contrast, AntennaFlow achieves precise alignment with the ground truth across the full angular spectrum with low ESS, demonstrating robustness to offsets in phaseless settings and preserving physically consistent radiation characteristics, including main lobes and sidelobes.

Table~\ref{tab:quantitative_comparison} further shows that AntennaFlow achieves the lowest RMS and Peak ESS on both datasets, indicating superior reconstruction quality with errors concentrated in less significant regions, while offering efficient inference much faster than TSWE and SRM and comparable to SET, demonstrating advantages in both fidelity and efficiency.

\begin{table}[htbp]
    \centering
    \caption{Quantitative Comparison of Far-Field Reconstruction Accuracy (RMS ESS and Peak ESS in dB)}
    \label{tab:quantitative_comparison}
    \renewcommand{\arraystretch}{1.2}
    \setlength{\tabcolsep}{4pt} 
    \begin{tabular}{cc|cccc}
        \hline
        Dataset & Metric & SET & TSWE &SRM &Ours \\
        \hline
        \multirow{2}*{MATLAB} & RMS ESS ($\downarrow$) & -18.97& -28.37 & -34.44 & \textbf{-36.90} \\
         & Peak ESS ($\downarrow$) & -6.91& -18.48 & -24.32 & \textbf{-27.06} \\
        \hline
        \multirow{2}*{FEKO} & RMS ESS ($\downarrow$) & -18.28 & -25.25 & -29.70 & \textbf{-33.81} \\
         & Peak ESS ($\downarrow$) & -8.42 & -15.34 & -18.33 & \textbf{-25.58}\\
        \hline
         - & Inference Time& \textbf{4s} & 10 min & 30 min & 9s \\
         \hline
    \end{tabular}

\end{table}

\subsubsection{Visualization Results}
To evaluate the reconstruction fidelity, we perform a comparative analysis of 
far-field (Fig.~\ref{fig:ff-r}) and near-field (Fig.~\ref{fig:2d_vis}) results. In the far field, our method demonstrates superior quality compared to SET, TSWE, and SRM. It corrects severe misalignment and reconstructs high-fidelity patterns from amplitude-only data while preserving main lobes and sidelobes. In the near field, Fig.~\ref{fig:2d_vis} groups multiple offset placements of the same antenna with their reconstructions and the ground truth. Despite diverse offsets, the reconstructions converge to nearly identical center-aligned fields and closely match the ground truth. This consistency under heterogeneous offset inputs is direct visual evidence that AntennaFlow extracts the offset-invariant embedding of the antenna and reconstructs its near-field amplitude, which is the property the downstream SET extraction relies on. 

\begin{table}[htbp]
    \centering
    \caption{Ablation Study on the Effect of Contrastive Pre-training}
    \label{tab:ablation}
    \renewcommand{\arraystretch}{1.2}
    \setlength{\tabcolsep}{6pt} 
    \begin{tabular}{c|cccc}
        \hline
        CL Use & NF RMS & NF Peak  & FF RMS & FF Peak \\
        \hline
        w/o CL & -25.74 & -19.18 & -24.43 & -16.46  \\
        w/ CL & -34.39 & -26.40 & -33.81 & -25.58  \\
        \hline
    \end{tabular}
\end{table}

\subsection{Validation of the Learned Embedding and Reconstruction}
The two points of AntennaFlow, an offset-invariant embedding and a deterministic flow-matching transport conditioned on it, together carry the burden of converting an offset amplitude into a center-aligned amplitude. We verify that each part behaves as intended through three complementary checks.

\begin{figure}[t]
    \centering
    \includegraphics[width=0.85\linewidth]{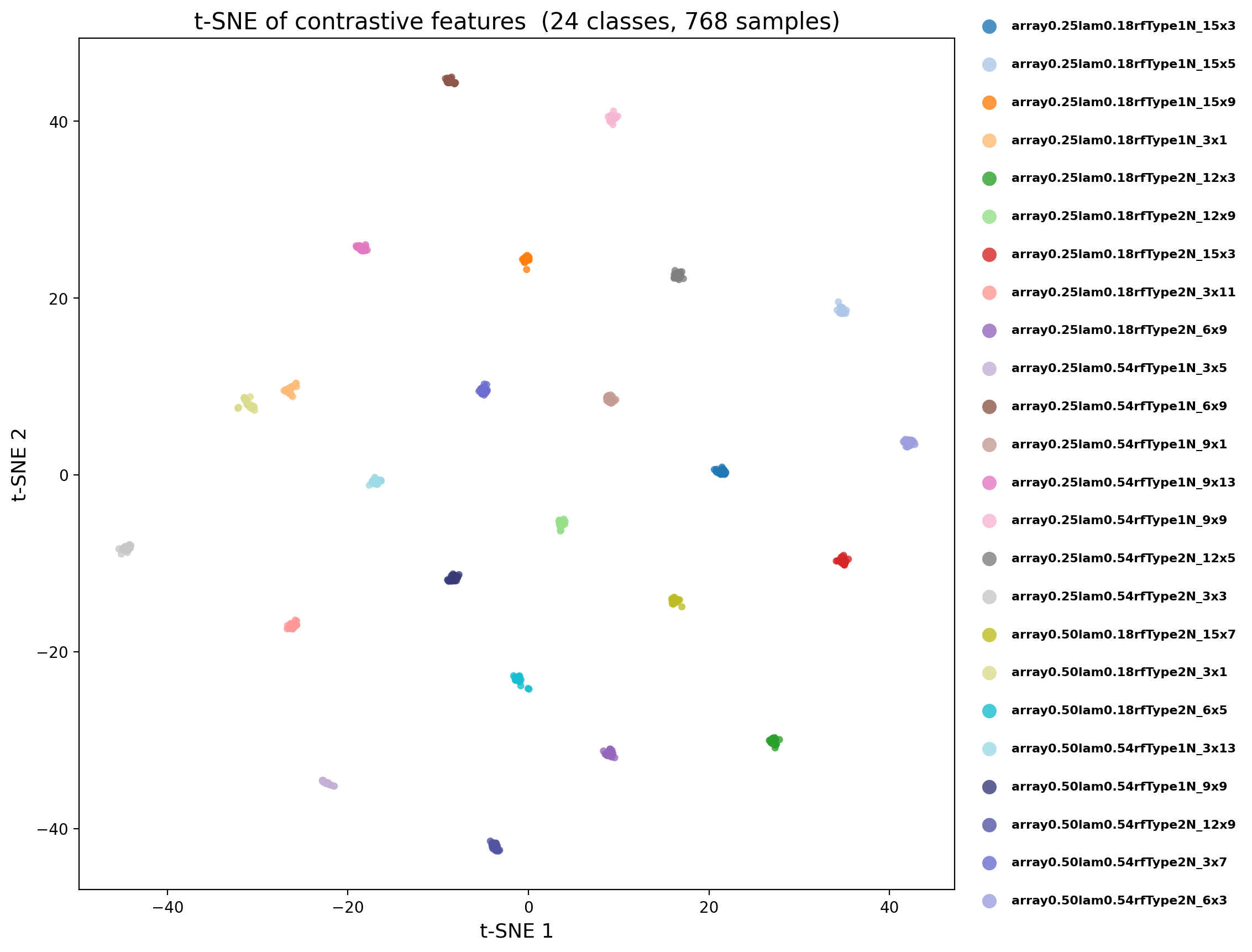}
    \caption{t-SNE projection of the encoder embeddings of $768$ amplitude samples drawn from $24$ distinct antenna configurations under random offsets. Colors denote antenna identity. Samples of the same antenna fall into tight clusters, while clusters of distinct antennas are clearly separated.}
    \label{fig:tsne}
\end{figure}

First, we verify that the encoder captures offset-invariant features by visualizing embeddings of offset samples from 24 antennas using t-SNE (Fig.~\ref{fig:tsne}). Embeddings from different antennas are well separated, while those from the same antenna under varying offsets form tight clusters, confirming that the encoder captures antenna-specific features invariant to offsets.

Second, we check whether this embedding suffices to drive the deterministic flow-matching transport back to the center-aligned field. As shown in Fig.~\ref{fig:2d_vis}, reconstructions from different offsets of the same antenna are nearly identical and closely match the ground truth. NF ESS and FF ESS evaluation further shows that the reconstructed fields preserve the key information while remaining physically consistent.

Third, we conduct an ablation study by removing contrastive pretraining and retraining the pipeline end-to-end. As shown in Table~\ref{tab:ablation}, all metrics degrade, confirming its key role in extracting the offset-invariant features for the framework.

\subsection{Sim-to-Real Robustness Evaluation}

To facilitate deployment in OTA antenna testing, we explicitly evaluate the sim-to-real gap arising from probe-dependent measurement responses and environmental noise. To approximate real measurement conditions, we introduce two representative perturbations: (1) Probe mismatch, using horn and dipole probes to simulate real-world sampling conditions and typically encountered in laboratory setups, and (2) introduced to model coupling effects and multipath interference arising from hardware imperfections and environmental uncertainty. These factors constitute the dominant sources of discrepancy between simulation and real OTA measurements.
As shown in Table~\ref{tab:robustness}, AntennaFlow remains stable under these perturbations, with only minor degradation in reconstruction accuracy, indicating strong robustness to realistic measurement conditions and effective generalization beyond idealized simulations.

\begin{table}[t]
\centering
\caption{Sim-to-Real Robustness Evaluation.(RMS ESS AND PEAK ESS IN DB)}
\label{tab:robustness}
\renewcommand{\arraystretch}{1.2}
    \setlength{\tabcolsep}{6pt} 
    \begin{tabular}{l|cccc}
        \hline
        Perturbation & NF RMS & NF Peak & FF RMS & FF Peak  \\
        \hline
        Default       & $-35.39$ & $-26.40$  & $-33.81$ & $-25.58$ \\
        \hline
        Horn Probe     & $-33.94$ & $-25.89$  & $-32.67$ & $-24.50$ \\
        \hline           
        Dipole Probe    &$-34.28$ & $-24.48$  & $-32.91$ & $-24.81$ \\
        \hline
        Additive noise   &$-33.49$ & $-25.18$  &  $-31.07$  & $-23.67$ \\
        \hline

    \end{tabular}
\end{table}

\section{Conclusion}
In this paper, we present AntennaFlow, which uses the basic measurement-frame fact that offset-mounted antenna measurements of the same antenna are merely different coordinate views of the same physical near field. A feature encoder turns this fact into an offset-invariant embedding directly extractable from offset-mounted amplitude measurements; a deterministic flow-matching transport conditioned on this embedding reconstructs the center-aligned near field; and the recovered center alignment is precisely the precondition required by SET to complete a phaseless, offset-vector-free NF--FF reconstruction. AntennaFlow outperforms SRM, the strongest baseline in our comparison, offering an accurate, efficient, and low-complexity solution for modern OTA antenna testing.

\bibliographystyle{IEEEtran}
\bibliography{globecom2026references}

\end{document}